\pdfoutput=1
\def\year{2020}\relax
\documentclass[letterpaper]{article} 
\usepackage{aaai20}  
\usepackage{times}  
\usepackage{helvet} 
\usepackage{courier}  
\usepackage[hyphens]{url}  
\usepackage{graphicx} 
\usepackage{graphicx}  
\usepackage{color,soul}
\usepackage{graphicx}
\usepackage{lipsum}
\usepackage{amsmath}
\usepackage{amssymb}
\usepackage{amsfonts}       
\usepackage{multirow}
\usepackage{booktabs}
\usepackage{nicefrac}
\usepackage{microtype}
\usepackage{cleveref}
\usepackage{bm}
\usepackage{bbm}
\usepackage{mathtools}
\usepackage{enumitem}
\usepackage{subcaption}
\usepackage{tikz}

\usepackage{algorithm}
\usepackage{algorithmic}
\newcommand{\citet}[1]{\citeauthor{#1} \shortcite{#1}}
\newcommand{\citep}{\cite}
\newcommand{\citealp}[1]{\citeauthor{#1} \citeyear{#1}}

\newcommand{\pkge}{HyperKG }
\newcommand{\transe}{TransE }
\newcommand{\rescal}{RESCAL }

\newcommand{\complex}{ComplEx }

\newcommand{\ftranse}{FTransE }
\newcommand{\dmult}{DISTMULT }

\newcommand{\vecc}[1]{\bm{#1}}

\newcommand{\vu}{\vecc{u}}
\newcommand{\vv}{\vecc{v}}
\newcommand{\ve}{\vecc{e}}
\newcommand{\vx}{\vecc{x}}
\newcommand{\vy}{\vecc{y}}
\newcommand{\vz}{\vecc{z}}

\newcommand{\vt}{\vecc{\theta}}
\newcommand{\he}{\vecc{s}}
\newcommand{\re}{\vecc{r}}
\newcommand{\te}{\vecc{o}}
\newcommand{\lossfn}{\mathcal{L}}
\newcommand{\regfn}{\mathcal{R}}

\newcommand{\manifold}[1]{\mathbb{#1}}
\newcommand{\tansp}{\mathcal{T}}

\newcommand{\R}{\mathbb{R}}

\newcommand{\tanspt}{\tansp_{\theta} \manifold{B}}

\newcommand{\retract}{\mathfrak{R}}
\DeclareMathOperator{\arccosh}{acosh}

\DeclareMathOperator*{\argmin}{arg\,min}

\newtheorem{lemma}{Lemma}
\newtheorem{proposition}{Proposition}

\title{HyperKG: Hyperbolic Knowledge Graph Embeddings for Knowledge Base Completion}
\author{Prodromos Kolyvakis\textsuperscript{\rm 1},  Alexandros Kalousis\textsuperscript{\rm 2}, Dimitris Kiritsis\textsuperscript{\rm 1}\\ 
\textsuperscript{\rm 1}\'Ecole Polytechnique F\'ed\'erale de Lausanne (EPFL), Lausanne, Switzerland, \\\textsuperscript{\rm 2}Business Informatics Department, University of Applied Sciences, \\
	Western Switzerland Carouge, HES-SO, Switzerland\\ 
}
 
\begin{document}

\maketitle

\begin{abstract}
Learning embeddings of entities and relations existing in knowledge bases allows the discovery of hidden patterns in data.
In this work, we examine the geometrical space's contribution to the  task of knowledge base completion.
We focus on the family of translational models, whose performance has been lagging, and propose a model, dubbed \textit{HyperKG}, which exploits the hyperbolic space in order to better reflect the topological properties of knowledge bases.
We investigate the type of regularities that our model can capture and we show that it is a prominent candidate for effectively representing a subset of Datalog rules.
We empirically show, using a variety of link prediction datasets, that hyperbolic space allows to narrow down significantly the performance gap between translational and bilinear models. 
\end{abstract}

\section{Introduction}

Learning in the presence of structured information is an important challenge for artificial intelligence \cite{muggleton1994inductive,richardson2006markov,getoor2007introduction}. 
Knowledge Bases (KBs) such as WordNet \cite{miller1998wordnet}, Freebase \cite{bollacker2008freebase}, YAGO \cite{suchanek2007yago} and DBpedia \cite{lehmann2015dbpedia} constitute valuable such resources needed for a plethora of practical applications, including question answering and information extraction. 
However, despite their formidable number of facts, it is widely accepted that their coverage is still far from being complete \cite{west2014knowledge}.

This shortcoming has opened the door for a number of studies addressing the problem of automatic knowledge base completion (KBC) or link prediction \cite{nickel2016review}. 
The impetus of these studies arises from the hypothesis that statistical regularities lay in KB facts, which when correctly exploited can result in the discovery of missing true facts \cite{P17-1088}.
Building on the great generalisation capability of distributed representations, a great line of research \cite{nickel2011three,bordes2013translating,Yang2015EmbeddingEA,nickel2016holographic,trouillon2016complex} has focused on learning KB vector space embeddings as a way of predicting the plausibility of a fact.

An intrinsic characteristic of knowledge graphs is that they present power-law (or scale-free) degree distributions as many other networks \cite{faloutsos1999power,steyvers2005large}. 
In an attempt of understanding scale-free networks' properties, various generative models have been proposed such as the models of \citet{barabasi1999emergence} and \citet{van2009random}.
Interestingly, \citet{PhysRevE.82.036106} have shown that scale-free networks naturally emerge in the hyperbolic space. 
Recently, the hyperbolic geometry was exploited in various works \cite{nickel2017poincare,Nickel2018LearningCH,Ganea2018HyperbolicEC,pmlr-v80-sala18a} as a means to provide high-quality embeddings for hierarchical structures.
Hyperbolic space has the potential to bring significant value in the task of KBC since it offers a natural way to take the KB's topological information into account.
Furthermore, many of the relations appearing in KBs lead to hierarchical and hierarchical-like structures \cite{Li:2016:HLP:2872518.2889387}.
 
At the same time, the expressiveness of various KB embedding models has been recently examined in terms of their ability to express any ground truth of facts \cite{kazemi2018simple,wang2018multi}. 
Moreover, \citet{gutierrez2018knowledge} have proceeded one step further and investigated the compatibility between ontological axioms and different types of KB embeddings. 
Specifically, the authors have proven that a certain family of rules, the quasi-chained rules which form a subset of Datalog rules \cite{abiteboul1995foundations}, can be exactly represented by a KB embedding model whose relations are modelled as convex regions; ensuring, thus, logical consistency in the facts induced by this KB embedding model.
In the light of this result, it seems important that the appropriateness of a KB embedding model should not only be measured in terms of fully expressiveness but also in terms of the rules that it can model.

In this paper, we examine whether building models that better reflect KBs' topological properties and rules brings performance improvements for KBC.
We focus on the family of translational models \cite{bordes2013translating} that attempt to model the statistical regularities as vector translations between entities' vector representations, and whose performance has been lagging.
We extend the translational models by learning embeddings of KB entities and relations in the Poincar\'{e}-ball model of hyperbolic geometry.
We do so by learning compositional vector representations \cite{P08-1028} of the entities appearing in a given fact based on translations.
The implausibility of a fact is measured in terms of the hyperbolic distance between the compositional vector representations of its entities and the learned relation vector.
We prove that the relation regions captured by our proposed model are convex.
Our model becomes, thus, a prominent candidate for representing effectively quasi-chained rules.

Among our contributions is the proposal of a novel KB embedding as well as a regularisation scheme on the Poincar\'{e}-ball model, whose effectiveness we prove empirically.
Furthermore, we prove that translational models do not suffer from the restrictions identified by \citet{kazemi2018simple} in the case where a fact is considered valid when its implausibility score is below a certain non-zero threshold.
Finally, we evaluate our approach on various benchmark datasets and our experimental results show that our approach makes a big step towards closing the performance gap between translational and bilinear models; demonstrating that the geometrical space's choice plays a significant role for KBC and illustrating the importance of taking into account both the topological and the formal properties of KBs.

%

\section{Related Work}

\textbf{Shallow KB Embedding Models.}
There has been a great line of research dedicated to the task of learning distributed representations for entities and relations in KBs.
To constrain the analysis, we only consider shallow embedding models that do not exploit deep neural networks or incorporate additional external information beyond the KB facts.
For an elaborated review of these techniques, please refer to \cite{nickel2016review,wang2017knowledge}.
We exclude from our comparison recent work that explores different types of training regimes, such as adversarial training, and/or the inclusion of reciprocal facts \cite{cai-wang-2018-kbgan,sun2018rotate,kazemi2018simple,pmlr-v80-lacroix18a} to make the analysis less biased to factors that could overshadow the importance of the geometrical space.

In general, the shallow embedding approaches can be divided into two main categories; the translational \cite{bordes2013translating} and the bilinear \cite{nickel2011three} family of models.
In the translational family, the vast majority of models \cite{wang2014knowledge,P15-1067,Xiao2016From,ebisu2018toruse} generalise \transe \cite{bordes2013translating}, which attempts to model relations as translation operations between the vector representations of the \emph{subject} and \emph{object} entities, as observed in a given fact.
In the bilinear family, most of the approaches \cite{Yang2015EmbeddingEA,nickel2016holographic,trouillon2016complex} generalise \rescal \cite{nickel2011three}, that proposes to model facts through bilinear operations over entity and relations vector representations.
In this paper, we focus on the family of translational models, whose performance has been lagging, and propose extensions in the hyperbolic space which by exploiting the topological and the formal properties of KBs bring significant performance improvements.

\textbf{Hyperbolic Embeddings.}
There has been a growing interest in embedding scale-free networks in the hyperbolic space \cite{boguna2010sustaining,papadopoulos2015network}.
The majority of these approaches are based on maximum likelihood estimation, that maximises the likelihood of the network's topology given the embedding model \cite{papadopoulos2015network}.
Additionally, hyperbolic geometry was exploited in various works as a way to exploit hierarchical information and learn more efficient representations.
Hyperbolic embeddings have been applied in a great variety of machine learning and NLP applications such as taxonomy reconstruction, link prediction, lexical entailment \cite{nickel2017poincare,Nickel2018LearningCH,Ganea2018HyperbolicEC,pmlr-v80-sala18a}.

Recently and in parallel to our work, \citet{balavzevic2019multi} studied the problem of embedding KBs in the hyperbolic space.
Similarly to our approach, the authors extend in the hyperbolic space the family of translational models demonstrating significant advancements over state-over-the-art.
However, the authors exploit both the hyperbolic as well as the Euclidean space by using the \emph{M\"obius Matrix-vector multiplication} and Euclidean scalar biases.
Unlike our experimental setup, the authors also include reciprocal facts.
Although their approach is beneficial, it becomes hard to quantify the contributions of hyperbolic space.
This is verified by the fact that their Euclidean model analogue performs in line with their ``hybrid'' hyperbolic-Euclidean model.
Finally, the authors do not study the expressiveness of their proposed model.

\section{Methods}

\subsection{Preliminaries}

We introduce some definitions and additional notation that we will use throughout the paper. 
We denote the vector concatenation operation by the symbol $\oplus$ and the inner product by $\langle \cdot,\cdot\rangle$.
We define the \emph{rectifier} activation function as: $[\cdot]_+ = \max(\cdot, 0)$.

\textbf{Quasi-chained Rules.} Let ${\bf E}, {\bf N}$ and ${\bf V}$ be disjoint sets of \emph{entities, (labelled) nulls} and \emph{variables}, respectively.\footnote{Only existential variables can be mapped to labelled nulls.}
Let ${\bf R}$ be the set of relation symbols.
A \emph{term} $t$ is an element in ${\bf E} \cup {\bf N} \cup{\bf V}$; an \emph{atom} $\alpha$ is an expression of the form $R(t_1, t_2)$, where $R$ is a \emph{relation} between the terms $t_1$, $t_2$. 
Let $\mathsf{terms}(\alpha) \vcentcolon= \{t_1, t_2\}$; $\mathsf{vars}(\alpha) \vcentcolon= \mathsf{terms}(\alpha) \cap {\bf V}$ and $B_n$ for $n \geq 0$, $H_k$ for $k \geq 1$ be atoms with terms in ${\bf E} \cup {\bf V}$. 
 Additionally, let $X_j \in \mathbf V$ for $j \geq 1$. 
 A \emph{quasi-chained (QC) rule} $\sigma$ \cite{gutierrez2018knowledge} is an expression of the form:
\begin{equation}\label{eqerule}
 B_1 \land \ldots \land B_n \rightarrow \exists X_1, \ldots ,X_j.H_1 \land \ldots \land H_k, 
 \end{equation}
 where for all $i:$ $1 \leq i \leq n$
$$|(\mathsf{vars}(B_1) \cup ... \cup \mathsf{vars}(B_{i-1})) \cap\mathsf{vars}(B_{i})| \leq 1$$

The QC rules constitute a subset of Datalog rules.
A \emph{database $D$} is a finite set of \emph{facts}, i.e., a set of \ atoms with terms in $\bf E$. A \emph{knowledge base (KB)} $\mathcal K$ constitutes of a pair $(\Sigma, D)$ where $\Sigma$ is an ontology whose axioms are QC rules and $D$ a database. 
It should be noted that no constraint is imposed on the number of available axioms in the ontology. 
The ontology could be minimal in the sense of only defining the relation symbols. 
However, any type of rule, whether it is the product of the ontological design or results from formalising a statistical regularity, should belong to the family of QC rules.
The Gene Ontology \cite{ashburner2000gene} constitutes one notable example of ontology that exhibits QC rules.

\textbf{Circular Permutation Matrices.}
An orthogonal matrix is a square real matrix whose columns and rows are orthogonal unit vectors (i.e., orthonormal vectors), i.e.
\begin{align}
Q^{\mathrm {T} }Q=QQ^{\mathrm {T} }=I
\end{align}
where I is the identity matrix. 
Orthogonal matrices preserve the vector inner product and, thus, they also preserve the Euclidean norms.
Let $1 \leq i < n$, we define the \emph{circular permutation matrix} $\Pi_i$ to be the orthogonal $n{\times}n$ matrix that is associated with the following circular permutation of a $n$-dimensional vector $\vecc{x}$:
\begin{align}
\left(\begin{matrix}
x_1 & \cdots & x_{n-i} & x_{n-i+1} & \cdots & x_n\\
x_{i+1} & \cdots & x_{n} & x_{1} & \cdots & x_{i}
\end{matrix}\right)
\end{align}
where $x_i$ is the ith coordinate of $\vecc{x}$ and $i$ controls the number of $n-i$ successive circular shifts.

\textbf{Hyperbolic Space.} Although multiple equivalent models of hyperbolic space exist, we will only focus on the Poincar\'{e}-ball model. 
The Poincar\'{e}-ball model is the
Riemannian manifold ${\manifold{P}^n = (\manifold{B}^n, d_p)}$, where
$\manifold{B}^n = \{\vx \in \R^n : \|\vx\| < 1\}$ and $d_p$ is the distance function \cite{nickel2017poincare}:
\begin{align}
d_p(\vu, \vv) & = \arccosh \left(1 + 2 \delta(\vu, \vv) \label{eq:pdist} \right) \\
\delta(\vu, \vv) & = \frac{\|\vu - \vv\|^2}{(1 - \|\vu\|^2)(1 - \|\vv\|^2)} \notag
\end{align}
The Poincar\'{e}-ball model presents a group-like structure when it is equipped with the \emph{M\"obius addition} \cite{ungar2012beyond}, defined by: 
\begin{align} \label{eq:mobius_add}
\vu +' \vv := \dfrac{(1+2 \langle \vu,\vv\rangle+\Vert \vv\Vert^2)\vu+(1-\Vert \vu\Vert^2)\vv}{1+2 \langle \vu,\vv\rangle+ \Vert \vu\Vert^2\Vert \vv\Vert^2} 
\end{align}
The isometries of $(\manifold{B}^n, d_p)$ can be expressed as a composition of a left gyrotranslation with an orthogonal transformation restricted to $\manifold{B}^n$, where the \textit{left gyrotranslation} is defined as $L_u: v \mapsto u+'v$ \cite{ahlfors1975invariant,rassias2019inequality}.
Therefore, circular permutations constitute zero-left gyrotranslation isometries of the Poincar\'{e}-ball model.

\subsection{\pkge}

The database of a KB consists of a set of facts in the form of $R(subject, object)$.
We will learn hyperbolic embeddings of entities and relations such that valid facts will have a lower implausibility score than the invalid ones.
To learn such representations, we extend the work of \citet{bordes2013translating}, and we define a translation-based model in the hyperbolic space; embedding, thus, both entities and relations in the same space. 

Let $\he, \re, \te \in \manifold{B}^{n}$ be the hyperbolic embeddings of the \emph{subject, relation and object}, respectively, appearing in the $R(subject, object)$ fact. 
We define a \emph{term embedding} as a function $\xi \colon \manifold{B}^{n}{\times}\manifold{B}^{n} \to \manifold{B}^{n}$, that creates a composite vector representation for the pair $(subject, object)$.
Since our motivation is to generalise the translation models to the hyperbolic space, a natural way to define the term embeddings is by using the M\"obius addition.
However, we found out empirically that the normal addition in the Euclidean space generalises better than the M\"obius addition.
To introduce non-commutativity in the term composition function, we use a circular permutation matrix to project the object embeddings.\footnote{The circular permutation operation retains the Euclidean norms.}
Non-commutativity is important because it allows to model asymmetric relations with compositional representations \cite{nickel2016holographic}.
Therefore, we define the term embedding as: $\he + \Pi_{\beta}\te$, where $\beta$ is a hyperparameter controlling the number of successive circular shifts.
To enforce the term embeddings to stay in the Poincar\'{e}-ball, we constrain all the entity embeddings to have a Euclidean norm less than $0.5$. 
Namely, $\|\ve\| < 0.5$ and $\|\re\| < 1.0$ for all entity and relation vectors, respectively.
The entities’ norm constraints do not restrict term embedding to span the Poincar\'{e}-ball.
We define the implausibility score as the hyperbolic distance between the term and the relation embeddings.
Specifically, the implausibility score of a fact is defined as:
\begin{align}
f_R(s, o) & = d_p(\he + \Pi_{\beta}\te, \re)
\end{align}
\begin{figure}
  \centering
    \includegraphics[width=0.8\columnwidth]{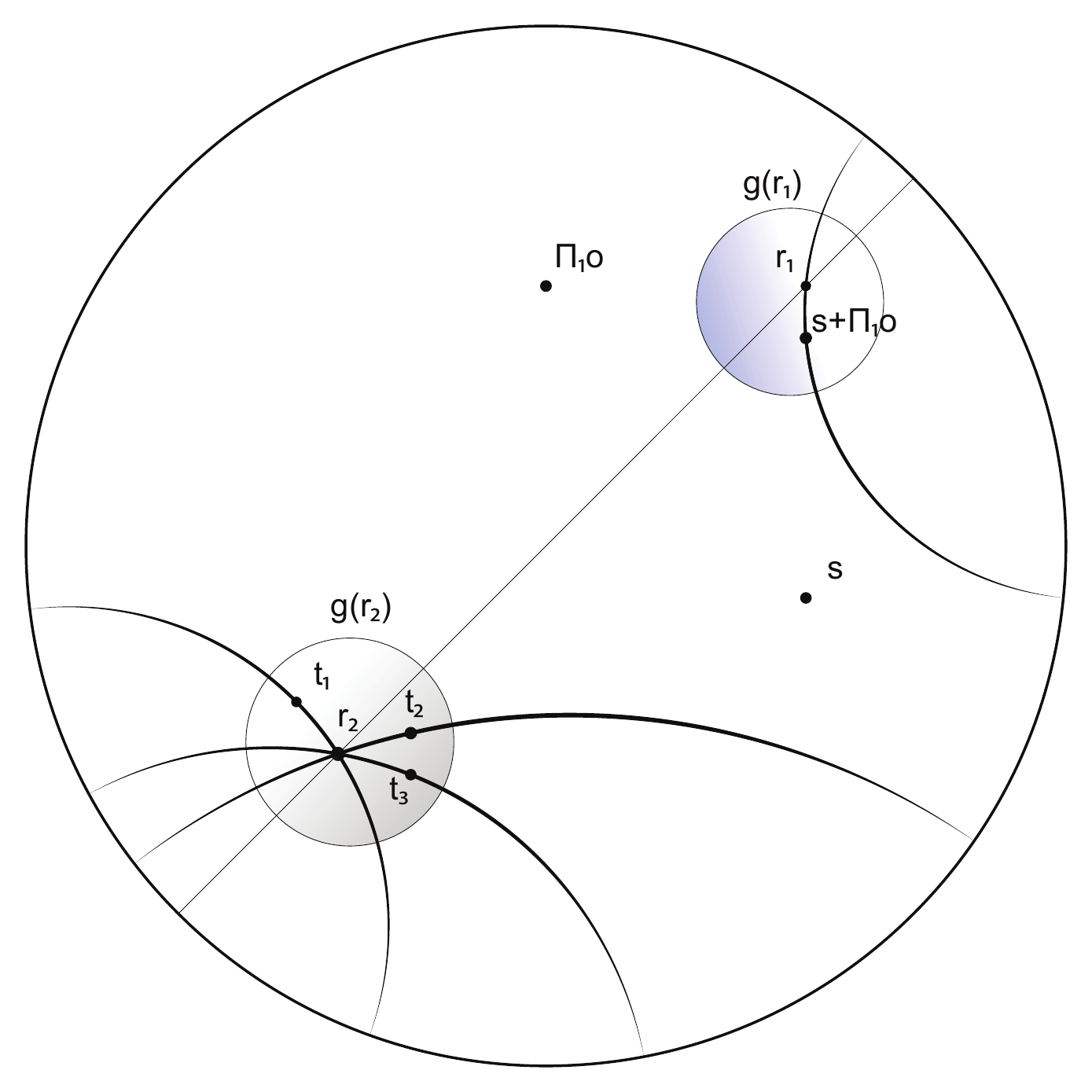}
    \caption{A visualisation of \pkge model in the $\manifold{P}^{2}$ space. The geodesics of the disk model are circles perpendicular to its boundary. The zero-curvature geodesic passing from the origin corresponds to the line $\epsilon: y-x=0$ in the Euclidean plane. Reflections over the line $\epsilon$ are equivalent to $\Pi_{1}$ permutations in the plane. $s, \Pi_{1} o, s+\Pi_{1} o$ are the subject vector, the permuted object vector and the composite term vector, respectively. $g(r_1), g(r_2)$ denote the geometric loci of term vectors satisfying relations $R_1, R_2$, with relation vectors $r1, r2$. $t_1, t_2, t_3$ are valid term vectors for the relation $R_2$.}
    \label{fig:poin disk}	
\end{figure}
\Cref{fig:poin disk} provides an illustration of the HyperKG model in $\manifold{P}^{2}$.
We follow previous work to minimise the following hinge loss function:
\begin{align}
\mathcal{L}=\sum_{\substack{R(s, o) \sim P,\\ R'(s', o') \sim N}} \left[ \gamma + f_{R}(s,o) -f_{R'}(s',o') \right]_+
\label{eq:hinge}
\end{align}
where $P$ is the training set consisting of valid facts, $N$ is a set of corrupted facts. 
To create the corrupted facts, we experimented with two strategies.
We replaced randomly either the subject or the object of a valid fact with a random entity (but not both at the same time).
We denote with $\#_{negs_{E}}$ the number of negative examples.
Furthermore, we experimented with replacing randomly the relation while retaining intact the entities of a valid fact.
We denote with $\#_{negs_{R}}$ the number of ``relation-corrupted'' negative examples.
We employ the {``\textit{Bernoulli}''} sampling method 
to generate incorrect facts \cite{wang2014knowledge,P15-1067,P17-1088}.

As pointed out in different studies \cite{bordes2013translating,DBLP:conf/aaai/DettmersMS018,pmlr-v80-lacroix18a}, regularisation techniques are really beneficial for the task of KBC.
Nonetheless, very few of the classical regularisation methods are directly applicable or easily generalisable in the Poincar\'{e}-ball model of hyperbolic space. 
For instance, the $\ell_2$ regularisation constraint imposes vectors to stay close to the origin, which can lead to underflows. 
The same holds for dropout \cite{JMLR:v15:srivastava14a}, when we used a rather large dropout rate.\footnote{In our experiments, we noticed that a rather small dropout rate had no effect on the model's generalisation capability.}
In our experiments, we noticed a tendency of the word vectors to stay close to the origin.
Imposing a constraint to the vectors to stay away from the origin stabilised the training procedure and increased the model's generalisation capability.
It should be noted that as the points in Poincar\'{e}-ball approach the ball's boundary their distance $d_p(\vu, \vv)$ approaches $d_p(\vu, \vecc{0}) + d_p(\vecc{0}, \vv)$, which is analogous to the fact that in a tree the shortest path between two siblings is the path through their parent \cite{pmlr-v80-sala18a}.
Building on this observation, our regulariser further imposes this ``tree-like'' property.
Additionally, since the volume in hyperbolic space grows exponentially, our regulariser implicitly penalises crowding.
Let $\Theta \vcentcolon= \{\ve_i\}_{i=1}^{|\bf E|} \bigcup \{\re_i\}_{i=1}^{|\bf R|} $ be the set of all entity and relation vectors, where $|\bf E|, |\bf R|$ denote the cardinalities of the sets $\bf E, R$, respectively. 
$\regfn(\Theta)$ defines the regularisation loss function that performed the best in our experiments:
\begin{equation}
	\regfn(\Theta) = \sum_{i=1}^{|\bf E| + |\bf R|}(1- \|\ \vecc{\theta}_i\|^2)
	\label{eq:reg scheme}
\end{equation}

The overall energy of the embedding is now defined as $\lossfn'(\Theta) = \lossfn(\Theta) + \lambda\regfn(\Theta)$, where $\lambda$ is a hyperparameter controlling the regularisation effect.
We define $a_i \vcentcolon= 0.5$, if $\theta_i$ corresponds to an entity vector and $a_i \vcentcolon= 1.0$, otherwise.
To minimise $\lossfn'(\Theta) $, we solve the following optimisation problem:
\begin{equation}
\Theta^\prime \gets \argmin_{\Theta} \lossfn'(\Theta) \quad\quad \text{s.t. } \forall\, \vt_i \in \Theta: \|\vt_i\| < a_i .\label{eq:loss}
\end{equation}
To solve \Cref{eq:loss}, we follow \citet{nickel2017poincare} and use Riemannian SGD (RSGD; \citealp{bonnabel2013stochastic}). 
In RSGD, the parameter updates are of the form:
\begin{equation*}
  \vt_{t+1} = \retract_{\vecc{\theta}_t}\left(-\eta \nabla_R \lossfn'(\vt_t) \right)
\end{equation*}
where \(\retract_{\vecc{\theta}_t}\) denotes the retraction onto the open $d$-dimensional unit ball at \(\vecc{\theta}_t\) and
\(\eta\) denotes the learning rate.
The Riemannian gradient of \(\lossfn'(\vt)\) is denoted by \({\nabla_R \in
\tanspt}\).
The Riemannian gradient can be computed as $\nabla_R = \frac{(1 - \|\vt_t\|^2)^2}{4} \nabla_E $, where $\nabla_E$
denotes the Euclidean gradient of \(\lossfn'(\vt)\). 
Similarly to \citet{nickel2017poincare}, we use the following retraction operation \(\retract_{\vecc{\theta}}(\vv) = \vt + \vv\).

To constrain the embeddings to remain within the Poincar\'{e} ball and respect the additional constraints, we use the following projection:
\begin{equation}
  \text{proj}(\vt, a) = \begin{cases}
  a\vt / (\|\vt\| + \varepsilon) & \text{if }\|\vt\| \geq a \\
  \vt & \text{otherwise ,}
  \end{cases}
  \label{eq:proj}
\end{equation}
where \(\varepsilon\) is a small constant to ensure numerical stability. 
In all experiments we used \(\varepsilon = 10^{-5}\).
Let $a$ be the constraint imposed on vector $\vt$, the full update for a single embedding is then of the form:
\begin{equation}
  \vt_{t+1} \gets \text{proj}\left(\vt_t - \eta \frac{(1 - \|\vt_t\|^2)^2}{4} \nabla_E , a\right) \label{eq:update} .
\end{equation}
We initialise the embeddings using the Xavier initialization scheme \cite{pmlr-v9-glorot10a}, where we use \Cref{eq:proj} for projecting the vectors whose norms violate the imposed constraints.

\begin{figure*}[!t]
     \centering
     \begin{subfigure}[b]{0.32\textwidth}
         \centering
         \includegraphics[width=\textwidth]{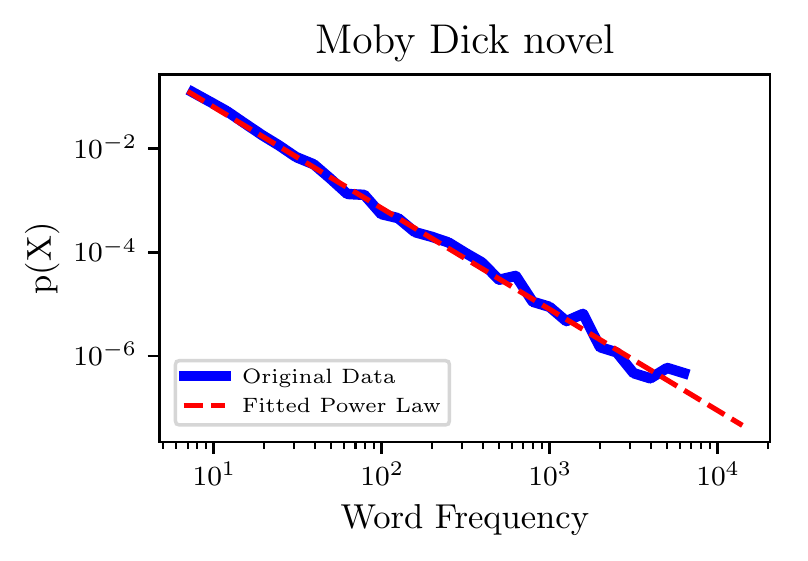}
         \label{fig:moby dick}
     \end{subfigure}
     \begin{subfigure}[b]{0.32\textwidth}
         \centering
         \includegraphics[width=\textwidth]{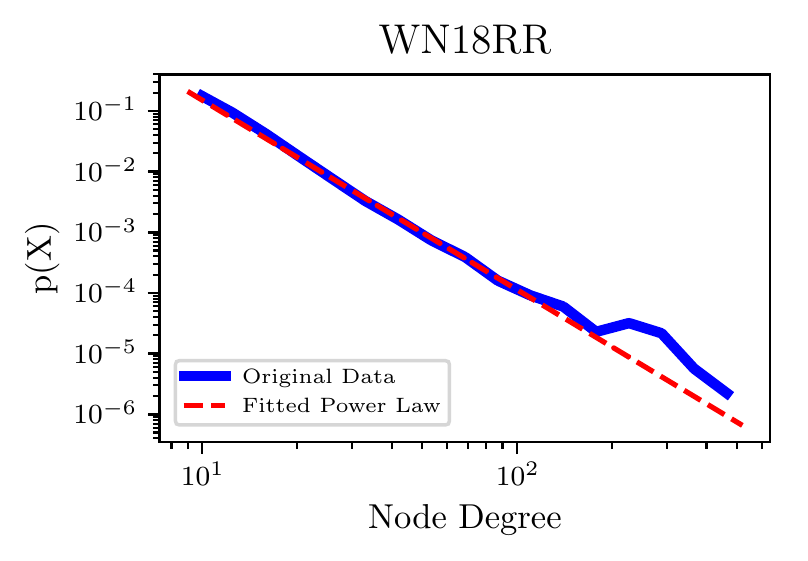}
         \label{fig:wn}
     \end{subfigure}
          \begin{subfigure}[b]{0.32\textwidth}
         \centering
         \includegraphics[width=\textwidth]{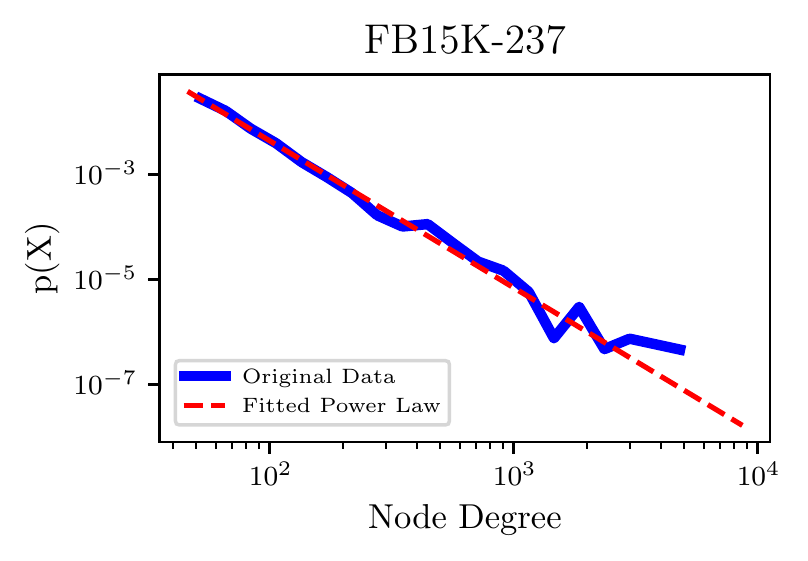}
         \label{fig:fb}
     \end{subfigure}
        \caption{ A visualisation of the probability density functions using a histogram with log-log axes.}
        \label{fig:power law}
\end{figure*}

\subsection{Convex Relation Spaces}

In this section, we investigate the type of rules that \pkge can model. 
Recently, \citet{wang2018multi} proved that the bilinear models are universal, i.e, they can represent every possible fact given that the dimensionality of the vectors is sufficient. 
The authors have also shown that the \transe model is not universal.
In parallel, \citet{kazemi2018simple} have shown that the \ftranse model \cite{Feng:2016:KGE:3032027.3032102}, which is the most general translational model proposed in the literature, imposes some severe restrictions on the types of relations the translational models can represent.
In the core of their proof lies the assumption that the energy function defined by the \ftranse model approaches zero for all given valid facts. 
Nonetheless, this condition can be considered less likely to be met from an optimisation perspective \cite{Xiao2016From}.

Additionally, \citet{gutierrez2018knowledge} studied the types of regularities that KB embedding methods can capture. 
To allow for a formal characterisation, the authors considered hard thresholds $\lambda_R$ such that a fact $R(s,o)$ is considered valid iff $s_R(\he,\te)\leq \lambda_R$, where $s_R(.,.)$ is the implausibility score. 
It should be highlighted that KB embeddings are often learned based on a maximum-margin loss function. 
Therefore, this assumption is not so restrictive. 
The vector space representation of a given relation $R$ can then be viewed as a region $n(R)$ in $\mathbb{R}^{2n}$, defined as follows:
\begin{align}
n(R) = \{\he \oplus \te \,|\, s_R(\he,\te) \leq \lambda_R\}
\label{eq:rel_regions}
\end{align}

Based on this view of the relation space, the authors prove that although bilinear models are fully expressive, they impose constraints on the type of rules that they can learn.
Specifically, let $R_1(X,Y) \rightarrow S(X, Y)$, $R_2(X,Y) \rightarrow S(X, Y)$ be two valid rules.
The bilinear models impose either that $R_1(X,Y) \rightarrow R_2(X,Y)$ or $R_2(X,Y) \rightarrow R_1(X,Y)$; introducing, thus, a number of restrictions on the type of subsumption hierarchies they can model.
\citet{gutierrez2018knowledge}, additionally, prove that there exists a KB embedding model with convex relation regions that can correctly represent knowledge bases whose axioms belong to the family of QC rules. 
Equivalently, any inductive reasoning made by the aforementioned KB embedding model would be logically consistent and deductively closed with respect to the ontological rules.
It can be easily verified that the relation regions of \transe \cite{bordes2013translating} are indeed convex.
This result is in accordance with the results of \citet{wang2018multi}; \transe is not fully expressive.
However, it could be a prominent candidate for representing in a consistent way QC rules.
Nonetheless, this result seems to be in conflict with the results of \citet{kazemi2018simple}.
Let $s_{R}^{TE}(s,o)$ be the implausibility score of TransE, we demystify this seeming inconsistency by proving the following lemma:
\begin{lemma}
	The restrictions proved by \citet{kazemi2018simple} can be lifted for the \transe model when we consider that a fact is valid iff $s_{R}^{TE}(s,o) \leq \lambda_R$ for sufficient $\lambda_R > 0$. 
	\label{le:trans}
\end{lemma}
We prove \Cref{le:trans} in the Appendix, by constructing counterexamples for each one of the restrictions. 
Since the restrictions can be lifted for the \transe model, we can safely conclude that they are not, in general, valid for all the generalisations of the \transe model.
In parallel, we built upon the formal characterisation of relations regions, defined in \Cref{eq:rel_regions} and we prove that the relation regions captured by \pkge are indeed convex. Specifically, we prove:
\begin{proposition}
The geometric locus of the term vectors, in the form of $\he + \Pi_{\beta}\te$, that satisfy the equation $d_p(\he + \Pi_{\beta}\te, \re) \leq \lambda_R$ for some $\lambda_R > 0$ corresponds to a $d$-dimensional closed ball in the Euclidean space. Let $\rho = \frac{\cosh(\lambda_R) - 1}{2}(1 - \|\re\|^2)$, the geometric locus can be written as $\|\he + \Pi_{\beta}\te - \frac{\re}{\rho + 1}\|^2 \leq \frac{\rho}{\rho + 1} + \frac{\|\re\|^2}{(\rho + 1)^2} -\frac{\|\re\|^2}{\rho + 1}$, where the ball's radius is guaranteed to be strictly greater than zero.
	\label{prop:convex}
\end{proposition}
The proof of \Cref{prop:convex} can be found in the Appendix.
By exploiting the triangle inequality, we can easily verify that the relation regions captured by \pkge are indeed convex.
\Cref{fig:poin disk} provides an illustration of the geometric loci captured by HyperKG in $\manifold{B}^{2}$.
This result shows that \pkge constitutes another one prominent embedding model for effectively representing QC rules.

\section{Experiments}

\subsection{Datasets}
We evaluate our \pkge model on the task of KBC using two sets of experiments. 
We conduct experiments on the WN18RR \cite{DBLP:conf/aaai/DettmersMS018} and FB15k-237 \cite{W15-4007} datasets.
We also construct two datasets whose statistical regularities can be expressed as QC rules to test our model's performance in their presence.
WN18RR and FB15k-237 constitute refined subsets of WN18 and FB15K that were introduced by \citet{bordes2013translating}. 
\citet{W15-4007} identified that WN18 and FB15K contained a lot of reversible relations, enabling, thus, various KB embedding models to generalise easily.
Exploiting this fact, \citet{DBLP:conf/aaai/DettmersMS018} obtained state-of-the-art results only by using a simple reversal rule.
WN18RR and FB15k-237 were carefully created to alleviate this leakage of information.

To test whether the scale-free distribution provides a reasonable means for modelling topological properties of knowledge graphs, we investigate the degree distributions of WN18RR and FB15k-237.
Similarly to \citet{steyvers2005large}, we treat the knowledge graphs as undirected networks.
We also compare against the distribution of the frequency of word usage in the English language; a phenomenon that is known to follow a power-law distribution \cite{zipf1949human}.
To do so, we used the frequency of word usage in Herman Melville’s novel ``Moby Dick'' \cite{newman2005power}.
We followed the procedure described by \citet{alstott2014powerlaw}.
In \Cref{fig:power law}, we show our analysis where we demonstrate on a histogram with log-log axes the probability density function with regard to the observed property for each dataset, including the fitted power-law distribution. 
It can be seen that the power-law distribution provides a reasonable means for also describing the degree distribution of KBs; justifying the work of \citet{steyvers2005large}.
The fluctuations in the cases of WN18RR and FB15k-237 could be explained by the fact that the datasets are subsets of more complete KBs; a fact that introduces noise which in turn can explain deviations from the perfection of a theoretical distribution \cite{alstott2014powerlaw}.

To test our model's performance on capturing QC rules, we extract from Wikidata \cite{vrandevcic2014wikidata,erxleben2014introducing} two subsets of facts that satisfy the following rules:
\begin{enumerate}[leftmargin=1cm,label=(\alph*)]
  \item $is\_a(\vx, \vy) \wedge part\_of(\vy, \vz) \rightarrow part\_of(\vx, \vz)$
  \item $part\_of(\vx, \vy) \wedge is\_a(\vy, \vz) \rightarrow part\_of(\vx, \vz)$
\end{enumerate}
Recent studies have noted that many real world KB relations have very few facts \cite{xiong-etal-2018-one}, raising the importance of generalising with limited number of facts.
To test our model in the presence of sparse long-tail relations, we kept the created datasets sufficiently small.
For each type of the aforementioned rules, we extract 200 facts that satisfy them from Wikidata.
We construct two datasets that we dub WD and WD$_{++}$.
The dataset WD contains only the facts that satisfy rule (\textbf{a}).
WD$_{++}$ extends WD by also including the facts satisfying rule (\textbf{b}).
The evaluation protocol was the following: For every dataset, we split all the facts randomly in train ($80\%$), validation ($10\%$), and test ($10\%$) set, such that the validation and test sets only contain a subset of the rules' consequents in the form of $part\_of(\vx, \vz)$.
\Cref{tab:datasets} provides details regarding the respective size of each dataset.
\begin{table}[!h]
\centering
\resizebox{7.5cm}{!}{
\setlength{\tabcolsep}{0.4em}
\begin{tabular}{l|rr|rrr}
\hline
\bf Dataset &  $\mid\bf E\mid$ & $\mid\bf R\mid$  & \bf $\#$Train & \bf $\#$Valid & \bf $\#$Test\\
\hline
WN18RR & 40,943 & 11 & 86,835 & 3,034 & 3,134\\
FB15k-237 & 14,541 & 237 & 272,115 & 17,535 & 20,466\\
WD & 418 & 2 & 550 & 25 & 25\\
WD$_{++}$ & 763 & 2 & 1,120 & 40 & 40\\
\hline
\end{tabular}
}
\caption{Statistics of the experimental datasets.}
\label{tab:datasets}
\end{table}  
\begin{table*}[!t]
\centering
\begin{tabular}{l|l|cc|cc}
\hline
\cline{3-6}
\multirow{2}{*}{\bf Method}& \multirow{2}{*}{\bf Type}&\multicolumn{2}{|c}{\bf WN18RR} & \multicolumn{2}{|c}{\bf FB15k-237}\\
\cline{3-6}
\cline{3-6}
&  & MRR & H@10 &   MRR  & H@10 \\
\hline
DISTMULT \citep{Yang2015EmbeddingEA} [$\star$] & Bilinear & 0.43 & 49 & 0.24 & 41\\
ComplEx \citep{trouillon2016complex} [$\star$] & Bilinear & 0.44 & 51 & 0.24 & 42\\
TransE \citep{bordes2013translating} [$\star$] & Translational & 0.22 & 50 & 0.29 & 46 \\
\hline
\pkge (M\"obius addition) & Translational & 0.30 & 44 & 0.19 & 32\\
\pkge (no regularisation) & Translational & 0.30 & 46 & 0.25 & 41\\
\pkge & Translational & 0.41 & 50 & 0.28 & 45\\
\hline
\end{tabular}
\caption{Experimental results on WN18RR and FB15k-237 test sets. MRR and H@10 denote the mean reciprocal rank and Hits@10 (in \%), respectively. [$\star$]: Results are taken from \citet{N18-2053}.  
} 
\label{tab:results}
\end{table*}
\subsection{Evaluation Protocol \& Implementation Details}
In the KBC task the models are evaluated based on their capability to answer queries such as $R(subject,\textbf{?})$ and $R(\textbf{?}, object)$ \cite{bordes2013translating}; predicting, thus, the missing entity.
Specifically, all the possible corruptions are obtained by replacing either the \emph{subject} or the \emph{object} and the entities are ranked based on the values of the implausibility score.
The models should assign lower implausibility scores to valid facts and higher scores to implausible ones.
We use the ``\textbf{Filtered}'' setting protocol \citep{bordes2013translating}, i.e., not taking any corrupted facts that exist in KB into account.
We employ two common evaluation metrics: mean reciprocal rank (MRR), and Hits@10 (i.e., the proportion of the valid test triples ranking in top 10 predictions). Higher MRR or higher Hits@10 indicate better performance.

The reported results are given for the best set of hyper-parameters evaluated on the validation set using grid search.
Varying the batch size had no effect on the performance.
Therefore, we divided every epoch into 10 mini-batches. 
The hyper-parameter search space was the following:
$\#_{negs_{E}} \in \{1, 2, 3, 4, 5, 8, 10, 12, 15\}$, $\#_{negs_{R}} \in \{0, 1, 2\}$, $\lambda \in \{2.0, 1.5, 1.0, 0.8, 0.6, 0.4, 0.2, 0.1, 0.0\}$, the embeddings' dimension $n \in \{40, 100, 200\}$, $\beta \in \{{\lfloor \frac{3n}{4} \rfloor}, {\lfloor \frac{n}{2}\rfloor}, {\lfloor \frac{n}{4} \rfloor}, 0\}$, $\eta \in \{0.8, 0.5, 0.2, 0.1, 0.05, 0.01, 0.005\}$ and $\gamma \in \{7.0, 5.0, 2.0, 1.5, 1.0, 0.8, 0.5, 0.2, 0.1\}$.
We used early stopping based on the validation's set filtered MRR performance, computed every 50 epochs with a maximum number of $2000$ epochs.

\subsection{Results \& Analysis}

\Cref{tab:results} compares the experimental results of our \pkge model with previous published results on WN18RR and FB15k-237 datasets.
We compare against the shallow KB embedding models \dmult \cite{Yang2015EmbeddingEA}, \complex \cite {trouillon2016complex} and \transe \citep{bordes2013translating}, which constitute important representatives of bilinear and translational models.
We exclude from our comparison recent work that explores different types of training regimes, such as adversarial training, the inclusion of reciprocal facts and/or multiple geometrical spaces \cite{cai-wang-2018-kbgan,sun2018rotate,kazemi2018simple,pmlr-v80-lacroix18a,balavzevic2019multi} to make the analysis less biased to factors that could overshadow the importance of the embedding space.
We give the results of our algorithm under the \pkge listing.

Despite the fact that \pkge belongs to the translational family of KB embedding models, it achieves comparable performance to the other models on the WN18RR dataset.
When we compare the performance of \pkge and TransE, we see that \pkge
achieves almost the double MRR score.
This consequently shows that the lower MRR performance of \transe is not an intrinsic characteristic of the translational models, but a restriction that can be lifted by the right choice of geometrical space.
With regard to Hits@10 on WN18RR, \pkge exhibits slightly lower performance compared to ComplEx.
On the FB15k-237 dataset, however, \pkge and \transe demonstrate almost the same behaviour outperforming \dmult and \complex in both metrics. 
Since the performance gap between \transe and \pkge is small, we hypothesise that it may be due to a less fine-grained hyperparameter tuning.

We also report in \Cref{tab:results} two additional experiments where we explore the behaviour of \pkge when the M\"obius addition is used instead of the Euclidean one as well as the performance boost that our regularisation scheme brings. 
In the experiment where the M\"obius addition was used, we removed the constraint for the entity vectors to have a norm less than $0.5$.
Although the M\"obius addition is non-commutative, we found beneficial to keep the permutation matrix. 
Nonetheless, we do not use our regularisation scheme.
Finally, the implausibility score is $d_p(\he +' \Pi_{\beta}\te, \re)$.
To investigate the effect of our proposed regularisation scheme, we show results where our regularisation scheme, defined in \Cref{eq:reg scheme}, is not used, keeping, however, the rest of the architecture the same. 
Comparing the performance of the \pkge variation using the M\"obius addition against the performance of the \pkge without regularisation, we can observe that we can achieve better results by using the Euclidean addition.
This can be explained as follows. 
Generally, there is no unique and universal geometrical space adequate for every KB \cite{gu2018learning}.
To recover Euclidean Space from the Poincar\'{e}-ball model equipped with the M\"{o}bius addition, the ball's radius should grow to infinity \cite{ungar2012beyond}.
Instead, by using the Euclidean addition and since the hyperbolic metric is locally Euclidean, \pkge can model facts for which the Euclidean Space is more appropriate by learning to retain small distances. 
Additionally, WN18RR contains more hierarchical relations compared to FB15k-237 \cite{balavzevic2019multi}, which further explains HyperKG’s performance boost on WN18RR.
Last but not least, we can observe that our proposed regularisation scheme is beneficial in terms of both MRR and Hits@10 on both datasets.

\Cref{tab:results rules} reports the results on the WD and WD$_{++}$ datasets.
We compare \pkge performance against \transe and ComplEx.
It can be observed that neither of the models manages to totally capture the statistical regularities of these datasets.
All the models present similar behaviour in terms of Hits@10.
\pkge and \transe, that both have convex relation spaces, outperform \complex on both datasets.
\pkge shows the best performance on WD, and demonstrates almost the same performance with \transe on WD$_{++}$.
\begin{table}[]
\centering
\begin{tabular}{l|cc|cc}
\hline
\cline{2-5}
\multirow{2}{*}{\bf Method}& \multicolumn{2}{|c}{\bf WD} & \multicolumn{2}{|c}{\bf WD$_{++}$}\\
\cline{2-5}
\cline{2-5}
 & MRR & H@10 &   MRR  & H@10 \\
\hline
ComplEx  & 0.92 & 98 & 0.81 & 92\\
TransE   & 0.88 & 96 & 0.89 & 98 \\
\hline
\pkge & 0.98 & 98 & 0.88 & 97\\
\hline
\end{tabular}
\caption{Experimental results on WD and WD$_{++}$ test sets. MRR and H@10 denote the mean reciprocal rank and Hits@10 (in \%), respectively.} 
\label{tab:results rules}
\end{table}
Our results point to a promising direction for developing less expressive KB embedding models which can, however, better represent certain rules.

\section{Conclusion and Outlook}
In this paper, we showed the geometrical space' significance for KBC by demonstrating that when models, whose performance has been lagging, are extended to the hyperbolic space, their performance increases significantly.
What is more, we demonstrated a new promising direction for developing models that better represent certain families of rules opening up for more fine-grained reasoning tasks.
Finally, recent hybrid models that exploit both the Euclidean and the hyperbolic space \cite{balavzevic2019multi} further demonstrate that hyperbolic space is a promising direction for KBC.
\section{Appendices}

\noindent\textbf{Proof of Lemma 1:} 
We begin by introducing the {\transe} model \cite{bordes2013translating}. 
In {\transe} model, the entities and the relations are represented as vectors in the Euclidean space.
Let, $\he, \re, \te \in \mathbb{R}^d$ denote the subject, relation and the object embedding, respectively. 
The implausibility score for a fact $R(s,o)$ is defined as $||\he+\re-\te||$.
Let $P$ define a set of valid facts. 
In the following we introduce some additional definitions needed for the introduction of the restrictions. 
\begin{itemize}
  \item A relation $r$ is \textbf{reflexive} on a set $E$ of entities if $(e,r,e) \in P$ for all entities $e\in E$. 
  \item A relation $r$ is \textbf{symmetric} on a set $E$ of entities if $(e_1,r,e_2)\in P \iff (e_2,r,e_1)\in P$ for all pairs of entities $e_1, e_2 \in E$. 
  \item A relation $r$ is \textbf{transitive} on a set $E$ of entities if $(e_1,r,e_2)\in P \wedge (e_2,r,e_3)\in P \Rightarrow (e_1,r,e_3)\in P$ for all $e_1,e_2,e_3\in E$. 
\end{itemize}
In the following, we list the restrictions mentioned in \citet{kazemi2018simple}.
\begin{itemize}
\item $\mathsf{R1:}$  If a relation $r$ is reflexive on $\Delta\subset E$, $r$ must also be symmetric on $\Delta$.
\end{itemize}

\begin{itemize}
\item $\mathsf{R2:}$  If $r$ is reflexive on $\Delta\subset E$, $r$ must also be transitive on $\Delta$.
\end{itemize}

\begin{itemize}
\item $\mathsf{R3:}$ If entity $e_1$ has relation $r$ with every entity in $\Delta\subset E$ and entity $e_2$ has relation $r$ with one of the entities in $\Delta$, then $e_2$ must have the relation $r$ with every entity in $\Delta$.
\end{itemize}

Let $n, m \in \mathbb{N}$, $i, j \in \mathbb{R}$ and $a \in \mathbb{R}_+^*$. 
Let $\vv = (v_1, v_2,\dots,v_m) \in \mathbb{R}^m$ and $\vu \in \mathbb{R}^n$.
We denote with $(v_1, v_2,\dots,v_m;\vu)$ the concatenation of vectors $\vv$ and $\vu$. Let $\vec{0}_n \in \mathbb{R}^n$ be the zero n-dimensional vector.  
For each restriction, we consider a minimum valid set of instances that could satisfy the restriction and we construct a counterexample that satisfies restriction's conditions but not the conclusion.
 
$\mathsf{ \bf R1:}$ This restriction translates to:
\begin{align}
\begin{rcases}
\|\vec{e_1}	+ \re - \vec{e_1}\| &\leq a \\
\|\vec{e_2}	+ \re - \vec{e_2}\| &\leq a \\
\|\vec{e_1}	+ \re - \vec{e_2}\| &\leq a \\
\end{rcases} 
\Rightarrow \|\vec{e_2}	+ \re - \vec{e_1}\| &\leq a
\end{align}
Let $n \geq 1$, $\re = (a;\vec{0}_{n-1})$, $\vec{e_1} = (i-a;\vec{0}_{n-1})$ and $\vec{e_2} = (i+a;\vec{0}_{n-1})$, then:
\begin{align}
\notag
\|\vec{e_2}	+ \re - \vec{e_1}\| &= \|((i+2a-(i-a));\vec{0}_{n-1})\| \Rightarrow \\
\|\vec{x_2}	 + \re- \vec{e_1}\| &= \sqrt{3}a > a
\end{align}

$\mathsf{ \bf R2:}$ This restriction translates to:
\begin{align}
\begin{rcases}
\|\vec{e_1}	+ \re - \vec{e_1}\| &\leq a \\
\|\vec{e_1}	+ \re - \vec{e_2}\| &\leq a \\
\|\vec{e_2}	+ \re - \vec{e_3}\| &\leq a \\
\|\vec{e_2}	+ \re - \vec{e_2}\| &\leq a \\
\|\vec{e_3}	+ \re - \vec{e_3}\| &\leq a \\
\end{rcases} 
\Rightarrow \|\vec{e_1}	+ \re - \vec{e_3}\| &\leq a
\end{align}
Let $n \geq 1$, $\re = (a;\vec{0}_{n-1})$, $\vec{e_1} = (i-a;\vec{0}_{n-1})$, $\vec{e_2} = (i+a;\vec{0}_{n-1})$ and $\vec{e_3} = (i+3a;\vec{0}_{n-1})$, then:
\begin{align}
\notag
\|\vec{e_1}	+ \re - \vec{e_3}\| &= \|((i-(i+3a);\vec{0}_{n-1})\| \Rightarrow \\
\|\vec{e_1}	+ \re - \vec{e_3}\| &= \sqrt{3}a > a
\end{align}

$\mathsf{ \bf R3:}$ This restriction translates to:
\begin{align}
\begin{rcases}
\|\vec{e_1}	+ \re - \vec{e_1}\| &\leq a \\
\|\vec{e_1}	+ \re - \vec{e_2}\| &\leq a \\
\|\vec{e_1}	+ \re - \vec{e_3}\| &\leq a \\
\|\vec{e_2}	+ \re - \vec{e_3}\| &\leq a \\
\end{rcases} \Rightarrow 
\begin{aligned}
\|\vec{e_2}	+ \re - \vec{e_2}\| &\leq a \\
\wedge \\
\|\vec{e_2}	+ \re - \vec{e_1}\| &\leq a
\end{aligned}
\end{align}
Let $n \geq 2$, $\re = (a;\vec{0}_{n-1})$, $\vec{e_1} = (i;\vec{0}_{n-1})$, $\vec{e_2} = (i+\frac{3a}{2},\frac{a}{2};\vec{0}_{n-2})$ and $\vec{e_3} = (i+2a;\vec{0}_{n-1})$, then:
\begin{align}
\notag
\|\vec{e_2}	+ \re - \vec{e_1}\| &= \|(i+\frac{3a}{2} +a -i,\frac{a}{2};\vec{0}_{n-2})\| \Rightarrow \\
\|\vec{e_2}	+ \re - \vec{e_1}\| &= \frac{\sqrt{26}}{2}a > a
\end{align}
It can be easily verified that these counterexamples also apply, with no modification, when the $\ell_1$ distance is used.
This ends our proof.

\noindent\textbf{Proof of Proposition 1}: Let $\|\he + \Pi_{\beta}\te\| < 1$, $\|\re\| < 1$ and $\lambda_R > 0$, we investigate the type of the geometric locus of the term vectors in the form of $\he + \Pi_{\beta}\te$ that satisfy the following equation:
\begin{align}
	d_p(\he + \Pi_{\beta}\te, \re) \leq \lambda_R
\end{align}
To simplify the notation, we denote $\vx := \he + \Pi_{\beta}\te$. 
\begin{align}
\notag
d_p(\vx, \re) &\leq \lambda_R &\iff  \\ \notag
1 + 2 \delta(\vx, \re) &\leq \cosh(\lambda_R) &\iff \\ 
\delta(\vx, \re) &\leq \frac{\cosh(\lambda_R) - 1}{2} 
\end{align}
Let $\alpha = (\cosh(\lambda_R) - 1)/2$. We should note that $\alpha > 0$, since $\forall x \in \mathbb{R^*}: cosh(x) > 1$. Then, we have:
\begin{align}
\frac{\|\vx - \re\|^2}{(1 - \|\vx\|^2)(1 - \|\re\|^2)} &\leq a 
\label{ineq:one}
\end{align}
Be setting $\rho = a(1 - \|\re\|^2)$, the inequality \ref{ineq:one} becomes:
\begin{align}
\notag
\|\vx - \re\|^2 \leq \rho(1 - \vx\|^2) &\iff  \\ \notag
(\rho + 1)\|\vx\|^2 - 2*\vx\re + \|\re\|^2 \leq \rho &\iff  \\ \notag
\|\vx\|^2 - 2*\vx\frac{\re}{\rho + 1} + \frac{\|\re\|^2}{\rho + 1} \leq \frac{\rho}{\rho + 1} &\iff  \\ 
\|\vx - \frac{\re}{\rho + 1}\|^2 \leq \frac{\rho}{\rho + 1} + \frac{\|\re\|^2}{(\rho + 1)^2} - \frac{\|\re\|^2}{\rho + 1}
\label{ineq:two}
\end{align}
We prove in the following that:
\begin{align}
\frac{\rho}{\rho + 1} + \frac{\|\re\|^2}{(\rho + 1)^2} - \frac{\|\re\|^2}{\rho + 1} > 0.
\end{align}
\label{ineq:three}
First, we note that since $\|\re\| < 1$, we also have that $\rho > 0$ based on the fact that $\alpha > 0$ and $1 - \|\re\|^2 > 0$.
Then, we have:
\begin{align}
	\notag
	& \frac{\rho}{\rho + 1} + \frac{\|\re\|^2}{(\rho + 1)^2} - \frac{\|\re\|^2}{\rho + 1} = \\ \notag
	&= \frac{1}{\rho + 1} \left( \rho + \frac{\|\re\|^2}{\rho + 1} - \|\re\|^2 \right) \\ \notag
	&= \frac{1}{\rho + 1} \left( \rho + \frac{1 - \rho-1}{\rho + 1}\|\re\|^2 \right) \\ \notag
	&= \frac{\rho}{\rho + 1} \left( 1 - \frac{1}{\rho + 1}\|\re\|^2 \right) 
	\label{eq:one}
\end{align}
We observe that $\frac{\rho}{\rho + 1} > 0$, hence, it is sufficient to check whether $1 - \frac{1}{\rho + 1}\|\re\|^2 > 0$.
We note that since $\|\re\| < 1$ and $\rho > 0$, we have $\frac{\|\re\|^2}{\rho+1} < \frac{1}{\rho+1}$.
However, $\frac{1}{\rho+1} < 1$. This concludes our proof.

\section{Models Parameters}

\subsection{\transe and \complex Implementation Details} 

For the experiments on the WD and WD$_{++}$ datasets, we used the public available implementations of \transe \cite{bordes2013translating} and \complex \cite{trouillon2016complex} provided in the OpenKE framework \cite{D18-2024}.
The reported results are given for the best set of hyper-parameters evaluated on the validation set using grid search.
We divided every epoch into 64 mini-batches. 

The hyper-parameter search space for \transe was the following: the dimensionality of embeddings $n \in \{50, 100\}$, SGD learning rate $\in \{0.0001, 0.0005, 0.001, 0.005\}$, $\mathit{l}_1$-norm or $\mathit{l}_2$-norm, and margin $\gamma \in \{1, 3, 5, 7\}$. 
The highest MRR scores  were achieved when using $\mathit{l}_1$-norm, learning rate at $0.005$, $\gamma$ = 7 and $n$ = 50 for both WD and WD$_{++}$.

The hyper-parameter search space for \complex was the following: $n \in \{50,100\}$, $\lambda \in \{0.1, 0.03, 0.01, 0.003, 0.001, 0.0003,0.0\}$, $\alpha_0 \in \{1.0, 0.5, 0.2, 0.1, 0.05, 0.02, 0.01\}$,  $\eta \in \{1, 2, 5, 10\}$ where $n$ the dimensionality of embeddings, $\lambda$ the $L^2$ regularisation parameter, $\alpha_0$ the AdaGrad's initial learning rate, and $\eta$ the number of negatives generated per positive training triple.
The highest MRR scores  were achieved when using learning rate at $0.05$, $\lambda$ = 0.1, $\eta$ = 5 and $n$ = 50 for WD.
For WD$_{++}$, the best hyper-parameters were achieved when using using learning rate at $0.05$, $\lambda$ = 0.1, $\eta$ = 5 and $n$ = 100.

\subsection{\pkge Parameters}
\begin{table*}[!t]
\centering
\begin{tabular}{l|l|cccccccc}
\hline
Dataset &  Model & $\#_{negs_{E}}$  & $\#_{negs_{R}}$ & $\eta$ & $\lambda$ & $n$ & $\gamma$, & $\beta$\\
\hline
WN18RR & \pkge & 10 & 0 & 0.01 & 0.8 & 100 & 1.0 & ${\lfloor \frac{n}{2}\rfloor}$\\
WN18RR & \pkge (M\"obius addition) & 10 & 0 & 0.01 & - & 100 & 1.0 & ${\lfloor \frac{n}{2}\rfloor}$\\
WN18RR & \pkge (no regularisation) & 10 & 0 & 0.01 & 0.0 & 100 & 1.0 & ${\lfloor \frac{n}{2}\rfloor}$\\
FB15k-237 & \pkge & 5 & 0 & 0.01 & 0.2 & 100 & 0.5 & ${\lfloor \frac{n}{2}\rfloor}$\\
FB15k-237 & \pkge (M\"obius addition) & 5 & 0 & 0.01 & - & 100 & 0.5 & ${\lfloor \frac{n}{2}\rfloor}$\\
FB15k-237 & \pkge (no regularisation) & 5 & 0 & 0.01 & 0.0 & 100 & 0.5 & ${\lfloor \frac{n}{2}\rfloor}$\\
WD & \pkge & 1 & 1 & 0.8 & 0 & 100 & 7 & ${\lfloor \frac{n}{2}\rfloor}$\\
WD$_{++}$  & \pkge & 1 & 1 & 0.1 & 0 & 100 & 7 & ${\lfloor \frac{n}{2}\rfloor}$\\
\hline
\end{tabular}
\caption{HyperKG's hyper-parameters used across the different experiments.}
\label{tab:used hyper parameters}
\end{table*}  
We report in the \Cref{tab:used hyper parameters} the best hyper-parameters for our \pkge model used across the different experiments.
For WD and WD$_{++}$, we do not use the {``\textit{Bernoulli}''} sampling method, but instead we corrupted the subject and object of a fact with equal probability. 


\begin{quote}
\fontsize{9.0pt}{10.0pt} \selectfont

\end{quote}

\end{document}